\documentclass[a4paper]{article}
\usepackage{multirow}
\usepackage{INTERSPEECH2016}
\usepackage{graphicx}
\usepackage{amssymb,amsmath,bm}
\usepackage{textcomp}
\usepackage{subfig}

\ninept

\title{Interactive Spoken Content Retrieval by Deep Reinforcement Learning}

\makeatletter
\def\name#1{\gdef\@name{#1\\}}
\makeatother \name{{\em Yen-Chen Wu$^1$, Tzu-Hsiang Lin$^2$, Yang-De Chen$^1$, Hung-Yi Lee$^1$, Lin-Shan Lee$^1$}}

\address{$^1$Graduate Institute of Communication Engineering, National Taiwan University \\
  $^2$Electrical Engineering, National Taiwan University \\
  {\footnotesize \tt \{yenchen.wu.wyc, iammrhelo, yangdechen0108, tlkagkb93901106\}@gmail.com, lslee@gate.sinica.edu.tw}
}

\begin{document}

\maketitle
\begin{abstract}
User-machine interaction is important for spoken content retrieval. For text content retrieval, the user can easily scan through and select on a list of retrieved item. This is impossible for spoken content retrieval, because the retrieved items are difficult to show on screen. Besides, due to the high degree of uncertainty for speech recognition, the retrieval results can be very noisy. One way to counter such difficulties is through user-machine interaction. The machine can take different actions to interact with the user to obtain better retrieval results before showing to the user. The suitable actions depend on the retrieval status, for example requesting for extra information from the user, returning a list of topics for user to select, etc. In our previous work, some hand-crafted states estimated from the present retrieval results are used to determine the proper actions. In this paper, we propose to use Deep-Q-Learning techniques instead to determine the machine actions for interactive spoken content retrieval. Deep-Q-Learning bypasses the need for estimation of the hand-crafted states, and directly determine the best action base on the present retrieval status even without any human knowledge. It is shown to achieve significantly better performance compared with the previous hand-crafted states.

  \end{abstract}


\vspace{-3mm}

  \section{Introduction}
	Interactive Information Retrieval (IIR)\cite{robins2000interactive,ruthven2008interactive} enhances a retrieval system by incorporating the user-system interaction into the retrieval process. The eventual goal of IIR is to guide the users to smoothly find out the desired information through the interactive process\cite{luo2014win,jin2013interactive}.
Previous interactive systems, for example, the  city guide \cite{misu2010bayes,misu2007speech} or the movie browser \cite{mcgraw2012automating,liu2012conversational}, usually had the content to be retrieved in text form stored in a semi-structured database, and made major efforts on transforming user’s natural language queries into semantic slots for subsequent database search.
    Here we focus on interactive retrieval of multimedia\cite{patton2002interactive,cho2004emotional} or spoken content\cite{garofolo2000trec}, which is radically different form text.

	There are several reasons which make user-system interaction important for spoken content retrieval. Firstly, speech recognition is still far from being perfect, and it inevitably produces errors which make the retrieval results uncontrollable. Moreover, the subword-based technologies are widely used in spoken content retrieval to deal with the OOV issue. These techniques result in higher recall rates, but also lead to lower precision rates.
    So very often many retrieved results are completely wrong. Also, it's difficult to show the retrieved multimedia or spoken information items on the screen, and it's hard for the users to scan through the retrieved results on the screen\cite{zhang2015information}.
    IIR is effective in updating the user instructions to boost the retrieval performance.

In several previous works~\cite{pan2012interactive,wen2012interactive,wen2013interactive}, Markov Decision Process (MDP) is used to model such spoken content IIR.
An earlier attempt is to let the user select among a list of retrieved key terms. However, this approach turned out to be inefficient for users.
A different approach was then proposed, in which the system has more actions to choose from and decides the most suitable action based on the present status including the present retrieved results, the number of interaction turns, and so on.
Some successful results have been achieved by first estimating some human-defined indicators from a set of features as the states and then selecting the actions based on the estimated states~\cite{wen2012interactive,wen2013interactive}.
However, such hand-crafted states may be inadequate for the purpose.
 And in those approaches, the state estimation and action selection are modeled as two cascading blocks and trained independently.
Without jointly estimating the whole process, both of them can be sub-optimal.

In this paper, we propose to use deep reinforcement learning in IIR for spoken content.
Deep reinforcement learning has recently achieved great renown and success, and deep-Q-Network (DQN) serves as a capable solution to learn from very raw inputs
~\cite{silver2016mastering, mnih2015human}.
In IIR with this approach, DQN can take the features originally used for state estimation as the input, and decide the actions directly.
Two major contributions were made in this paper.
First, we show that the hand-crafted states used previously, like average precision, cannot represent the retrieval status very well, and cascading the estimation of such states with action selection based on such states is definitely not optimal. This is why the DQN proposed here, which is an end-to-end approach, achieved remarkable improvements.
Secondly, through utilizing DQN, we show that even using the raw feature--the retrieval scores from the search engine--as the DQN input outperformed the previous approach with hand-crafted states.

\begin{figure*}[t]
        \centering
        \includegraphics[width=\linewidth]{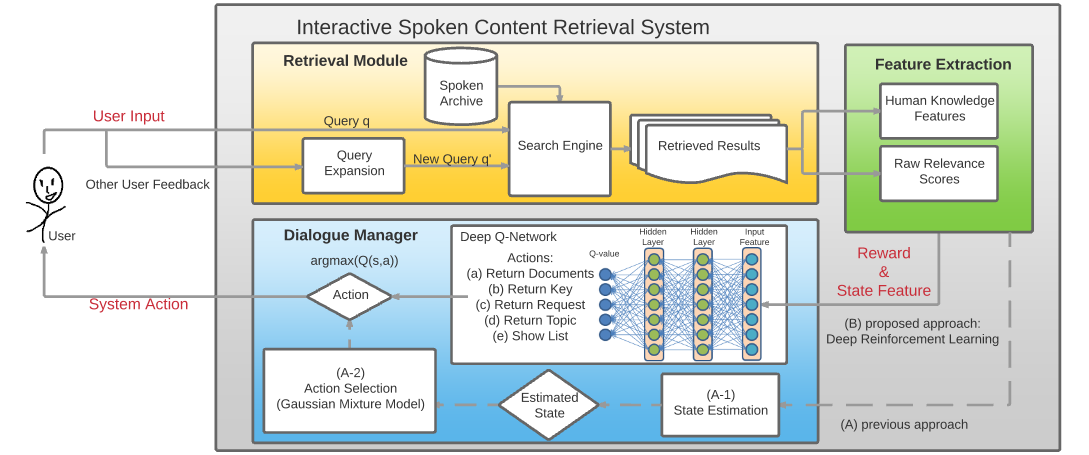}
        \caption{{\it Block diagram of the proposed approach compared to the previous approach in Path (A) (dash line)}}
        \vspace{-5mm}
        \label{fig:diagram}
\end{figure*}

\vspace{-3mm}
\section{Proposed Approach}
The framework for the proposed approach is depicted in Fig.~\ref{fig:diagram}.
At the left hand side, the user first enters a query $q$ into the system.
With the user query $q$ or plus other feedback information from the user during the interaction, the retrieval module (in the upper middle) described in Section \ref{subsec:retrieval} will generate a list of retrieved result.

The system takes actions based on the retrieved results to interact with the user as discussed in Section \ref{subsec:action}.
A set of features extracted from the retrieved results (at the upper right corner), and the dialogue manager (in the lower middle) determining the action based on the extracted feature is introduced in Section \ref{subsec:feature} and 2.4 respectively.
In the dialogue manager, there are paths (A) and (B) for determine the actions.
Path (A) (in Section 2.4.1) is the previous approach with estimated states, while path (B) (in Section 2.4.2) uses the deep reinforcement learning proposed in this paper.

\subsection{Retrieval Module} \label{subsec:retrieval}

\subsubsection{Language Modeling Retrieval Module}
The basic idea of a language model based retrieval framework is to represent the query $q$ and a document $d$ both as language models \(\theta_q\) and $\theta_d$.
More details about estimating $\theta_d$ for spoken documents are left out here ~\cite{lafferty2001document,chia2010statistical}.
The relevance score $S(q,d)$ for the given query $q$ and a document $d$, which is used to rank the documents $d$ during the retrieval process, is evaluated based on the KL divergence  between \(\theta_q\) and \(\theta_d\), or $S(q,d) = -KL(\theta_q \| \theta_d)$.
Furthermore, the user may designate a set of terms to be irrelevant to what he/she is seeking for, which can be modeled as a negative information model $\theta_N$.
Thus, the complete relevance score $S(q,d)$ considers both the query model $\theta_q$ and the negative information model $\theta_N$ as below.

\vspace{-5mm}
\begin{equation}
S(q,d) = -[KL(\theta_q \| \theta_d) - \beta KL(\theta_N \| \theta_d)],
\label{eq2}
\end{equation}
\vspace{-2mm}
where $\beta$ as an adjustable parameter~\cite{chia2010statistical,lee2012improved}.

\subsubsection{Query-Regularized Mixture Model for Query Expansion}
After receiving the feedback from the user during the iteration, the system generates a new query model $\theta'_q$.
We adopt the query-regularized mixture model~\cite{lee2012improved,zhai2001model,tao2006regularized} previously proposed for pseudo-relevance feedback to estimate the new query models $\theta'_q$.
However, in this query expansion process, the new query model may be tainted by unrelated information in the pseudo-relevant documents.
We therefore regularize the query models using a key term set, which is initially composed of the original query and can be expanded throughout the interactive session.

\subsection{Actions to be taken by the system} \label{subsec:action}
In order to help the user offer useful information, with which the system can retrieve documents better matched to the user's goal, five actions are defined for user-system interaction as presented below.

(a) \textit{Return Documents}: The dialogue manager returns the current list of retrieved results ranked by $S^k(q, d)$ at the present time $k$ in decreasing order and asks the user to select a relevant document.

(b) \textit{Return Key Term}: the dialogue manager asks the user whether a key term $t^{\ast}$ is relevant.

(c) \textit{Return Request}: the dialogue manager asks the user to provide an additional query term $\hat t$.

(d) \textit{Return Topic}: The dialogue manager returns a list of topics generated with latent topic models \cite{hofmann1999probabilistic,blei2003latent,blei2006dynamic,rosen2004author} and asks the user to select one.

(e) \textit{Show list}: The dialogue manager shows the retrieved results ranked by $S^k(q, d)$ to the user and ends the interactive session.

With actions (a),(b),(c),(d) the system receives extra information from the user and generates a new query $q'$ accordingly for the next step retrieval. Action (e) ends the interactive session and shows the retrieved results to the user.

\subsection{Feature Extraction} \label{subsec:feature}
Sets of features describing the characteristics or present status of the retrieved results from the search engine in Section~\ref{subsec:retrieval} is extracted, based on which the proper actions are selected in Section 2.4.
Two sets of features are tested here:
\begin{itemize}
\item \textbf{Human Knowledge Feature}: A set of features hand-crafted based on human knowledge is extracted. These features were used in the previous works \cite{wen2013interactive}. Examples for these features include clarity score \cite{he2006query}, query scope \cite{he2006query}, the simplified query clarity score (SCS) \cite{he2006query}, ambiguity score \cite{cronen2002predicting}, similarity between the query and the collection \cite{zhao2008effective}, weighted information gain (WIG) \cite{zhou2007query} and query feedback \cite{zhou2007query}.
\item \textbf{Raw Relevance Scores}: Considering the power of deep learning, the dialogue manager may be able to make decisions simply based on the raw relevance scores of retrieved items without any human knowledge.
Here the relevance scores of the top-$N$ documents in the retrieved results are taken as the features with $N$-dimensions.
\end{itemize}

\subsection{Dialogue Manager}

The dialogue manager is based on Markov Decision Process (MDP).
MDP~\cite{bellman1957markovian} is defined as a tuple $\{\mathcal{S}, \mathcal{A}, \mathcal{T}, \mathcal{R}, \gamma \}$, where $\mathcal{S}$ is the set of states, $\mathcal{A}$ the set of actions, $\mathcal{T}(s'|s,a)$ is the transition probability of ending up in state $s'$ when executing action $a$ in  state $s$, $\mathcal{R}$ is the reward function, and $\gamma$ the discount factor. A mapping from a state $s \in S$ to an action $a \in A$, or action selection at each state, is a policy $\pi$. Given a policy $\pi$, the value of the Q-function ${(Q^\pi : S \times A \to R)}$ is the estimation of the expected  discounted sum of all rewards that can be received over an infinite state transition path starting from state s taking action $\pi(s)$:
${Q^π (s,a) = E[ \sum^{\infty}_{k=0} \gamma^k r_k|s_0 = s, a_0 = a]}$,
where $r_k$ is the reward received from the action $a_k$ taken at state \(s_k\), and k is the sequence index for states and actions.
The optimal policy maximizes the value of each state-action pair: ${\pi^{\ast} (s) = arg\max_{a\in \mathcal{A}} Q^{\ast} (s, a)}$,
so finding an optimal policy is equivalent to finding the optimal Q-function.


The reward of $a_k$,the action taken at state $s_k$, is defined as
\vspace{-1mm}
\begin{equation}
r_k = -C_k + \tau [E(s_k) - E(s_{k-1})].
\vspace{-2mm}
\label{eq_reward}
\end{equation}
$E(s)$ is some retrieval quantity metric at the state $s$ and $C_k$ is the estimated effort by the user to perform the action $a_k$.
$\tau$ is a trade-off parameter between user effort and the retrieval quality improvement.

\subsubsection{Previous Approach: Estimating the hand-crafted state} \label{subsec:previous} 

The previous approach~\cite{wen2012interactive,wen2013interactive} for the dialogue management is path (A) in Figure \ref{fig:diagram}.
The underlying assumption behind this approach is that the proper choice of the action can be made by considering some evaluation metric for the retrieved results, which is the average precision (AP) here.
This assumption leads to a two-stage process, shown as blocks A-1 and A-2, in the dialogue manager of Fig.\ref{fig:diagram}.
Block A-1 is the state estimation.
It takes the feature set extracted from the retrieved results in Section \ref{subsec:feature} as the input, and estimates the AP for them, taken as the state $s$ in the $Q$-function for action selection.
Block A-2 is for action decision.
It uses fitted value iteration (FVI)~\cite{chandramohanoptimizing} to train a Gaussian mixture model (GMM) to approximate the $Q$-value function $Q(s,a)$ for each action $a$, and the system takes the action $a$ with the maximum $Q(s,a)$.

This approach have several weaknesses.
First, the evaluation metric AP is not necessary a good representation for states.
AP simply indicates how well the retrieved results are.
Empirical results have shown that even with the same AP, the optimal actions can be different.
However, it's not easy to come up with better state definition with human knowledge.
Next, it's not able to fix the error margin for relatively weak state estimation, because the state estimation (block A-1) and action selection (block A-2) are separately trained rather than considered jointly.

\subsubsection{Proposed Approach: Deep Reinforcement Learning} \label{subsec:deep}
The proposed approach used Deep-Q-Network (DQN) to do deep reinforcement learning of $Q$-function.
DQN is able to overcome the problems of the previous approach mentioned above at least to some degree. As can be  seem in path (B) of Fig~\ref{fig:diagram}, the DQN directly generates the proper action from the input features through the hidden layers. In this way the error propagation for the two cascaded stages (block A-1 and block A-2 in path (A)) is eliminated, and the machine automatically learns from the  features extracted from the retrieve module including the human knowledge features and the raw relevance scores of the retrieved results.

The DQN is a deep neural network (DNN)~\cite{hinton2006reducing,bengio2009learning} with parameters $\theta$ to estimate the state-action value function $Q(s,a;\theta)$\footnote{Because the state-action value function $Q(s,a)$ here depends on the deep neural network parameters $\theta$, the function should be written as $Q(s,a;\theta)$.}.
The input of the DQN is the features extracted in Section \ref{subsec:feature}, while its output  dimension  is the same as the number of possible actions $a$, and the output is the state-action value $Q(s, a; \theta)$ for each action $a$ in the action set $\mathcal{A}$.
The DQN is trained by iteratively updating the parameters $\theta$.
With parameters obtained at the $i$-th iteration, denoted as $\theta^i$, $\theta$ can be learned by minimizing the following the loss function $L_i(\theta^i)$ in (\ref{eq7}) by gradient descent.
\vspace{-1mm}
\begin{equation}
L_i(\theta^i) = \mathop{\mathbb{E}}_{s,a,r,s' \sim \mathcal{U(\mathcal{D})} } [( \hat{y}_i - Q(s,a;\theta^i))^2].
\vspace{-2mm}
\label{eq7}
\end{equation}
$\mathcal{D} = \{e_1, e_2, ..., e_t,...e_L\}$ includes experiences $e_t = (s_t, a_t, r_t, s_{t+1})$ (taking action $a_t$ at state $s_t$ obtaining reward $r_t$ and reaching state $s_{t+1}$ at the next time step) is a dataset collected from many retrieval episodes to be used for training.
The expression $s,a,r,s' \sim \mathcal{U(\mathcal{D})}$ in (\ref{eq7}) means, instead of using the current experience as prescribed by the standard temporal-difference learning, the network is trained by sampling mini-batches of experiences from $\mathcal{D}$ uniformly at random.
This method is referred to as experience replay, which is a key ingredient behind the success of DQN.
In this way the efficiency in using the training data can be improve through re-use of the experience samples in multiple updates, and the correlation among the samples used in the update can be reduced through the uniform sampling from the replay buffer~\cite{lin1993reinforcement,mnih2015human}.

$\hat{y}_i$ in (\ref{eq7}) is defined as below:
\vspace{-1mm}
\begin{equation}
\hat{y}_i = r+\gamma \cdot \max_{a' \in \mathcal{A}} Q(s',a';\theta^-)
\vspace{-2mm}
\label{eq8}
\end{equation}
where $\theta^{-}$ represents the parameters of a fixed and separate target network, which is taken here as the parameters obtained several iterations before.
Freezing the parameters of the target network $Q(s', a'; \theta^-)$ for a fixed number of iterations while updating the online network $Q(s, a; \theta^i)$ is another key innovation for the success of DQN~\cite{lin1993reinforcement,mnih2015human}, which improved the stability of the algorithm.

\vspace{-3mm}
\section{Experiments}
\subsection{Experiment Setting}
We used a broadcast news corpus in Mandarin Chinese recorded from radio or TV stations in Taipei from 2001 to 2003 as the target document archive to be retrieved. There was a total of 5047 news documents with a total length of 198 hours. We used one-best transcriptions and lattices for the spoken archive. We used a tri-gram language model trained on 39M words of Yahoo news. 163 text queries and their relevant spoken documents (not necessarily including the query terms) were provided by 22 graduate students.
We used DQN with two and four hidden layers of 1024 nodes and relu as the activation function~\cite{maas2013rectifier}. The DQN framework is modified from an open-source code~\footnote{https://github.com/spragunr/deep\_q\_rl}.
Mean Average Precision (MAP) was selected as our retrieval evaluation metric. The costs of actions were set empirically considering the extra burden caused when the user provides feedback. And 10-fold cross validation was performed in all experiments.

We generated simulated users with the following behavior for training the dialogue manager.
\begin{itemize}
\item(a) \textit{Return Documents}: The simulated user viewed the list from the top and chooses the first relevant document.
\item(b) \textit{Return Key Term}: Replied "YES" if the key term appeared in more than 50\% of the relevant documents and "NO" otherwise.
\item(c) \textit{Return Request}: Entered a key term based on Tf-idf (term frequency-inverse document frequency).
\item(d) \textit{Return Topic}: Randomly returned one of the relevant topics manually labeled by graduate students.
\end{itemize}


\begin{figure}[th]
        \centering
        \includegraphics[width=\linewidth]{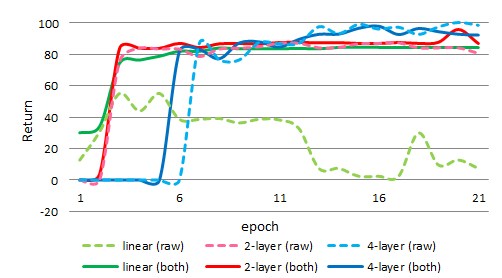}
        \caption{{\it Learning curves with either raw relevance scores alone (for top-100 retrieved items) (dotted curves) or with human knowledge  features in addition (solid curves) with different DQN depths over lattice transcriptions.}}
        \label{fig:model}
\end{figure}

\begin{figure}[t]
        \centering
        \includegraphics[width=\linewidth]{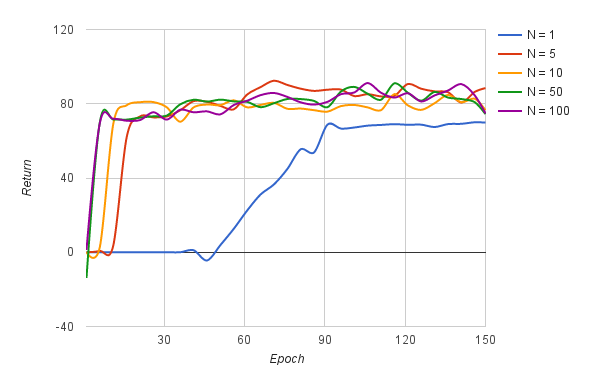}
        \caption{{\it Learning curves of DQN with 2 layers using different sizes of relevance scores.  N is the number of the top-N retrieved items which raw relevance scores were used.}}
        \label{fig:retrievalscores2}
\end{figure}

\begin{table}[t]
\caption{\label{tab:result} {\it MAP and Return for (a) previous approach, (b) proposed approach and (c) upper bound evaluated on both one-best transcriptions and lattices.}}
\vspace{2mm}
\centerline{
\begin{tabular}{ l | c|c|c|c }
\hline
\multicolumn{1}{c|}{Approaches} &
\multicolumn{2}{c}{one-best} &
\multicolumn{2}{c}{lattices} \\
&  MAP & Return &  MAP &  Return \\
\hline \hline
(a-1) First-pass  &0.4521 & - & 0.4577 & - \\
(a-2) Random Actions & 0.4553 & -61.7 & 0.4117 & -111.21 \\
(a-3) Hand-crafted States & 0.5398 & 67.07 & 0.5626 & 84.54 \\
\hline
(b-1) Raw Feature & 0.5619 & 89.27 & 0.5847  & 105.03  \\
(b-2) Selected Feature+(b-1) & 0.5691 & 95.72 & 0.5907  & 110.90  \\
\hline
(c) Upper Bound (Oracle) & 0.6554 & 164.94 & 0.6639 & 168.75  \\
\hline
\end{tabular}
}
\end{table}

\subsection{Result and Discussion}
Table \ref{tab:result} shows the results in MAP and Return $R=\sum_{k=0}^{T} r_k$ achieved on spoken content retrieval module based on either one-best transcriptions (left half) or lattices (right half).

$\bullet$ (a) Baselines: (a-1) are the first-pass results without any interaction, and their returns can be regarded as 0.
(a-2) show the results for taking random actions, repeating 1000 times to estimate the MAP and expected return, and (a-3) are the results of the previous work with state estimation.

$\bullet$ (b) DQN with different features:
(b-1) are the results when only the raw relevance scores were used.
And (b-2) combine the human knowledge features and the raw relevance scores in (b-1).

$\bullet$ (c) Oracle: Obtained by brute-force search over every action sequence whose length was less than 5, and picked the action sequence with the highest return.
It is considered as the upper bound for the finite interactive scenario.

From (a-2), we notice that taking actions randomly is not a good way to improve the retrieval process: although the system gains extra information after every interaction, inefficient requests from the system may also burden the user and yield poor returns.
The previous hand-crafted state approach in (a-3) obtained much better performance than random actions in (a-2) in terms of both MAP and return.
Surprisingly, the proposed approach in (b-1) using DQN end-to-end reinforcement learning only with raw relevance scores without any human knowledge easily outperformed the results using plenty of human knowledge by estimating hand-crafted states in (a-3).
This shows that the end-to-end neural network can properly estimate useful action selection indicators implicitly in its hidden layers directly from raw retrieval scores.
Finally, using both the raw relevance scores and human knowledge features together in (b-2) achieved the best results.

To further explore DQN's ability on selecting actions,
Figure~\ref{fig:model} shows the learning curves of the final return for the proposed approach with different neural network depth using different sets of features: raw relevance scores for the top 100 retrieved items (dotted curves) and plus the human knowledge features (solid curves). Colors green, red, blue represents the curves using linear model (neural network without hidden layers), 2 and 4 layers, respectively.
We observe that the linear model can be easily trained when the two sets of features are jointly used.
But when solely using raw relevance scores, it diverged and didn't work (dotted-green).  This shows that the mapping from raw relevance scores to proper actions is complicated and cannot be modeled without any hidden layers.
The red and blue curves for 2 and 4 hidden layers showed that deeper model usually yielded better performance in most cases, especially when using raw relevance scores exclusively.  But with a deeper model, it took more epochs to converge.  This figure shows the results experimented on the lattice transcription, whereas the results on one-best transcription were similar.

Since the results using raw relevance scores alone is comparable to those using both sets of features jointly with DQN, we conducted an extra experiment using raw relevance scores alone but with different sizes on the one-best transcription.
Over DQN of 2 layers, Figure 3 shows the learning curves with the raw relevance scores alone, where N is the number of the retrieved items which relevance scores were used. We observe that DQN could learn from only 1 relevance score, though it gave a relatively poor performance, while all others (5,10,50,100) converged to some good return.


\section{Conclusion}

Due to the high degree of uncertainty in speech and the difficulty of showing on screen to users,  user-system interaction is highly desired for spoken content retrieval.
In this paper, we utilize Deep-Q-network(DQN) to learn better state-action values without estimating the hand-crafted states which were used previously.  This end-to-end learning is able offer overall optimization for user-system interaction and produce significant improvements on return.
We further found that even with raw relevance scores alone without any human knowledge, we achieved very good performance.  We hope the results seen here can light up the future directions of DQN in interactive spoken language systems.

\newpage
\eightpt

\bibliographystyle{IEEEtran}
\bibliography{mybib}

\end{document}